%% file: main.tex
\documentclass[10pt,twocolumn,letterpaper]{article}

\usepackage[pagenumbers]{cvpr}      
\usepackage{algorithm}
\usepackage{algorithmic}
\usepackage{multirow}
\usepackage{pifont} 

\definecolor{cvprblue}{rgb}{0.21,0.49,0.74}
\usepackage[pagebackref,breaklinks,colorlinks,allcolors=cvprblue]{hyperref}
\usepackage{algorithm}

\def\confName{CVPR}
\def\confYear{2025}

\title{GATE3D: Generalized Attention-based Task-synergized Estimation in 3D\thanks{This paper was accepted (Poster) to the 3rd Workshop on Computer Vision for Mixed Reality (CV4MR) at CVPR 2025.}}
\author{%
Eunsoo Im%
\quad Changhyun Jee 
\quad Jung Kwon Lee
\\[2pt]
Superb AI\\
{\tt\small \{eslim,chjee,jklee\}@superb‑ai.com}%
}

\begin{document}
\maketitle
\input{sec/0_abstract}    
\input{sec/1_intro}

\input{sec/2_relative_works}
\input{sec/3_method}

\input{sec/4_experiments}
\input{sec/5_conclusion}

{
    \small
    \bibliographystyle{ieeenat_fullname}
    \bibliography{main}
}
\input{sec/X_suppl}


\end{document}

%% file: sec/0_abstract.tex
\begin{abstract}
The emerging trend in computer vision emphasizes developing universal models capable of simultaneously addressing multiple diverse tasks. Such universality typically requires joint training across multi-domain datasets to ensure effective generalization. However, monocular 3D object detection presents unique challenges in multi-domain training due to the scarcity of datasets annotated with accurate 3D ground-truth labels, especially beyond typical road-based autonomous driving contexts.
To address this challenge, we introduce a novel weakly supervised framework leveraging pseudo-labels. Current pretrained models often struggle to accurately detect pedestrians in non-road environments due to inherent dataset biases. Unlike generalized image-based 2D object detection models, achieving similar generalization in monocular 3D detection remains largely unexplored.
In this paper, we propose GATE3D, a novel framework designed specifically for generalized monocular 3D object detection via weak supervision. GATE3D effectively bridges domain gaps by employing consistency losses between 2D and 3D predictions. Remarkably, our model achieves competitive performance on the KITTI benchmark as well as on an indoor-office dataset collected by us to evaluate the generalization capabilities of our framework.
Our results demonstrate that GATE3D significantly accelerates learning from limited annotated data through effective pre-training strategies, highlighting substantial potential for broader impacts in robotics, augmented reality, and virtual reality applications.
Project page: \url{https://ies0411.github.io/GATE3D/}
\end{abstract}

%% file: sec/1_intro.tex
\section{Introduction}
\label{sec:intro}

\begin{figure}[t]
    \centering
    \includegraphics[width=\columnwidth]{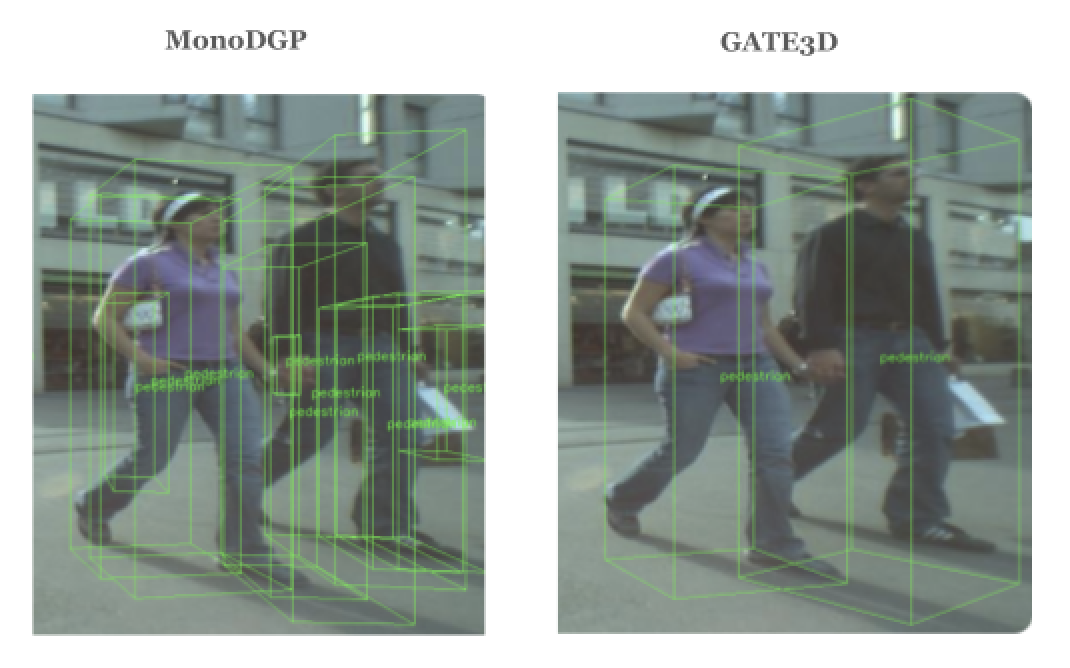}  
    \caption{Qualitative results of Query Diversity}  
    \label{Fig:query}  
\end{figure}

Predicting objects in 3D space from single images remains a critical yet challenging task with wide-ranging implications across robotics, surveillance, augmented reality (AR), and virtual reality (VR). Recent advances in 2D object detection, driven by abundant annotated datasets, have produced highly generalized models. However, real-world applications frequently involve understanding spatial relationships in 3D environments, making accurate 3D object detection indispensable \cite{wang2024revisiting}.
Existing monocular 3D object detection methods predominantly leverage publicly available autonomous driving datasets such as KITTI\cite{geiger2012kitti}, NuScenes\cite{caesar2020nuscenes}, and Waymo\cite{sun2020waymo}, which offer precise 3D ground truth annotations. However, these datasets are heavily biased toward road-based scenarios, significantly restricting model generalization to non-road contexts. As a result, models pretrained solely on these datasets often fail to detect pedestrians effectively in indoor settings like offices or commercial environments, as these scenarios differ substantially in viewpoint and object distribution.
To bridge this dataset gap, we propose a novel weakly supervised approach. We collected a diverse dataset(Synergy3D) consisting of 100k images captured from various sites, including offices, malls, and public squares. Unlike traditional methods, which typically require extensive use of additional sensors (e.g., LiDAR) and labor-intensive manual annotation processes, our framework efficiently generates pseudo-3D labels by integrating existing 2D detectors, orientation predictors, and depth estimators \cite{wang2024training}. Our innovative pseudo-labeling pipeline significantly accelerates the labeling process, enhancing efficiency by up to 100 times compared to conventional manual annotation approaches. The inherent imperfections of pseudo-labels are effectively mitigated by the proposed weakly supervised training strategy.
Based on this comprehensive dataset, we introduce GATE3D, a novel monocular 3D detection framework featuring attention-based architectures specifically tailored for improved generalization across diverse domains \cite{zhang2024heightformer}. Our model consistently outperforms existing monocular 3D detection approaches, demonstrating substantial robustness across varied scenarios.
Recognizing the diversity of camera sensors in our supplemental dataset, which leads to inconsistencies in resolution and intrinsic parameters, we propose a novel normalization method leveraging virtual-space transformations. This method effectively standardizes focal lengths and image resolutions, significantly mitigating domain discrepancies and further enhancing our model's robustness.

Our main contributions can be summarized as follows

\begin{itemize}
\item We introduce a large-scale, diverse, pseudo-labeled 3D object detection dataset created efficiently without reliance on LiDAR or intensive manual labeling.
\item We propose GATE3D, an attention-based monocular 3D detection framework designed explicitly for enhanced generalization across multiple scenarios.
\item We develop a novel virtual-space normalization technique addressing sensor diversity, further improving robustness and generalization performance.
\end{itemize}

%% file: sec/2_relative_works.tex
\section{Related Work}
\label{sec:related_work}
\subsection{Monocular 3D Object Detection}
Monocular 3D detection, predicting accurate 3D bounding boxes from single images, is inherently challenging due to limited depth cues. Approaches based solely on single images typically leverage deep learning architectures without auxiliary sensors. For example, M3D-RPN \cite{brazil2019m3d} introduces a 3D region proposal network with depth-wise convolution, SMOKE \cite{liu2020smoke} adopts a CenterNet-based approach, and MonoFlex \cite{zhang2021objects} utilizes edge heatmaps to enhance detection of truncated objects. MonoPair \cite{chen2020monopair} exploits spatial relationships between objects, and MonoCon \cite{liu2022learning} employs additional monocular contexts to boost inference performance.

In contrast, methods utilizing auxiliary data such as LiDAR or depth maps generally achieve higher accuracy through richer spatial information. Examples include ROI-10D's \cite{manhardt2019roi} dense depth maps, CaDDN's \cite{reading2021categorical} bird’s-eye-view representations, and CMKD’s \cite{gong2022cmkd} knowledge distillation from LiDAR. Despite accuracy gains, these methods face practical limitations due to sensor complexity and resource demands.

Unlike previous works, our GATE3D approach does not rely on additional sensors or manual annotations, significantly enhancing labeling efficiency through foundational pseudo-labeling.

\subsection{Transformer-based Methods}
Recent methods increasingly adopt Transformer architectures due to their capacity for global context aggregation and dynamic feature representation. MonoDETR and MonoDGP integrates depth prediction with Transformer-based decoding \cite{zhang2022monodetr,pu2024monodgp}, while MonoDTR \cite{huang2022monodtr} employs depth-aware positional encodings for enhanced depth reasoning.

\subsection{Unified Datasets and Generalization}
Current 3D datasets (KITTI \cite{geiger2012kitti}, NuScenes \cite{caesar2020nuscenes}, Waymo \cite{sun2020waymo}) primarily target outdoor driving scenarios, limiting model generalization. OMNI3D \cite{brazil2023omni3d} attempts dataset unification to broaden domain coverage. Building upon these efforts, we present a comprehensive pseudo-labeled dataset encompassing indoor and varied environments, alongside our novel virtual-space normalization technique to mitigate sensor biases, ultimately improving the generalization capabilities of GATE3D.

%% file: sec/3_method.tex
\section{Method}
\label{sec:method}

\begin{figure*}[t]
    \centering
    \includegraphics[width=1.0\textwidth]{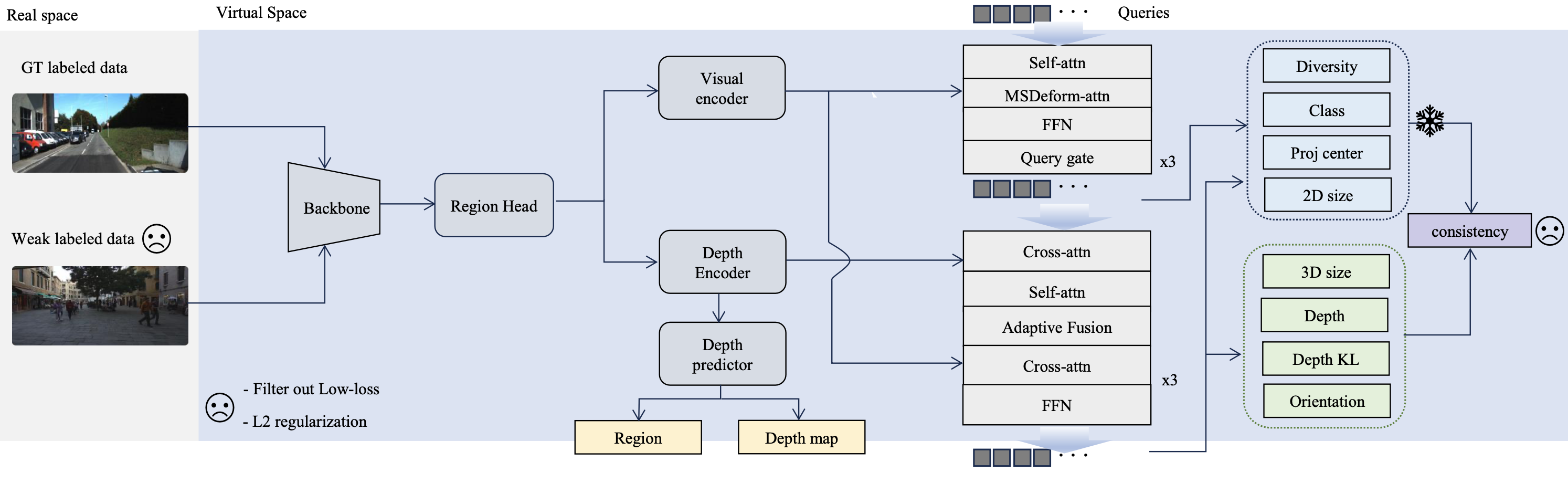}  
    \caption{GATE3D architecture overview. The proposed framework incorporates a DETR-style 3D detection backbone enhanced with attention-based modules, and supports both fully and weakly supervised learning modes. For ground-truth-labeled samples, the model is trained via standard 3D detection losses. For weakly labeled data, pseudo-3D annotations are generated from 2D detection, monocular depth estimation, and orientation prediction. To mitigate label noise, we introduce a 2D–3D consistency loss that aligns projected 3D box dimensions with frozen 2D predictions. Notably, during weak supervision, the 2D detector remains fixed to preserve its reliability, while only the 3D decoder is optimized. This hybrid learning strategy improves robustness and generalization across diverse domains.}  
    \label{Fig:overview}  
\end{figure*}

\subsection{GATE3D}
\noindent \textbf{Query Gate and Adaptive Fusion Module}
We introduce a novel Query Gate and Adaptive Fusion Module designed to enhance the decoder architecture of existing DETR-based mono-camera 3D detectors \cite{zhang2022monodetr, pu2024monodgp}. The Query Gate employs a learned, sigmoid-activated linear layer to dynamically scale individual query vectors \cite{li2024gafusion}, selectively emphasizing essential information while suppressing irrelevant details. This adaptive scaling ensures that the model effectively retains salient features during the decoding process.

The Adaptive Fusion Module addresses the limitation of conventional methods, which typically consider only the query-key relationships and neglect direct utilization of depth information \cite{zhou2021fusionpainting}. Even depth-aware attention approaches generally rely on static depth positional embeddings, leading to an inability to dynamically adapt query responses to varying depth contexts. To overcome this, we propose extracting the global context from depth positional embeddings and dynamically integrating it with query features \cite{samfusion2024}. A linear layer followed by a sigmoid activation function is utilized to modulate the influence of depth features adaptively for each query \cite{iet2024aviff}.
Formally, given the query feature vector \(\mathbf{tgt}_i \in \mathbb{R}^{d_{model}}\) and the global depth context vector \(\mathbf{g} \in \mathbb{R}^{d_{model}}\), the adaptive gating is computed as:
\begin{equation}
\mathbf{gate}_i = \sigma\left(\mathbf{W}\left[\mathbf{tgt}_i; \mathbf{g}\right]\right),
\end{equation}
where \(\mathbf{W} \in \mathbb{R}^{d_{model}\times 2d_{model}}\) denotes a learnable linear transformation and \(\sigma\) denotes the sigmoid activation function. The gated output is then obtained by element-wise multiplication:
\begin{equation}
\mathbf{tgt}_i^{\text{fused}} = \mathbf{gate}_i \odot \mathbf{tgt}_i
\end{equation}

\noindent \textbf{Query Diversity}
Typically, query-based detectors do not require Non-Maximum Suppression (NMS). However, we observed that previous transformer-based mono-camera 3D detectors \cite{zhang2022monodetr, pu2024monodgp} frequently produce duplicate object predictions, necessitating the use of NMS. Our analysis suggests that excessive cosine similarity between queries from the final 3D decoder contributes to redundant and unstable predictions. To address this issue, we propose a diversity-promoting loss term designed to minimize the average cosine similarity among queries, thereby enhancing prediction diversity and stability \cite{antoniussen2024visled}. Fig. \ref{Fig:query} demonstrates the effectiveness of our proposed query diversity approach.

\begin{equation}
    L_{\text{diversity}} = \frac{1}{B Q (Q - 1)} \sum_{b=1}^{B} \sum_{i=1}^{Q} \sum_{\substack{j=1 \\ j \ne i}}^{Q} \frac{q_{b,i} \cdot q_{b,j}}{\|q_{b,i}\|\|q_{b,j}\|}
\end{equation}
where $B$ is the batch size, $Q$ is the number of queries, and $q_{b,i}$ represents the embedding vector of the $i$-th query in batch $b$.
Fig. \ref{Fig:query} illustrates the effectiveness of our proposed query diversity loss, confirming its capability to significantly reduce redundancy and improve prediction stability.

\begin{figure}[t]
    \centering
    \includegraphics[width=\columnwidth]{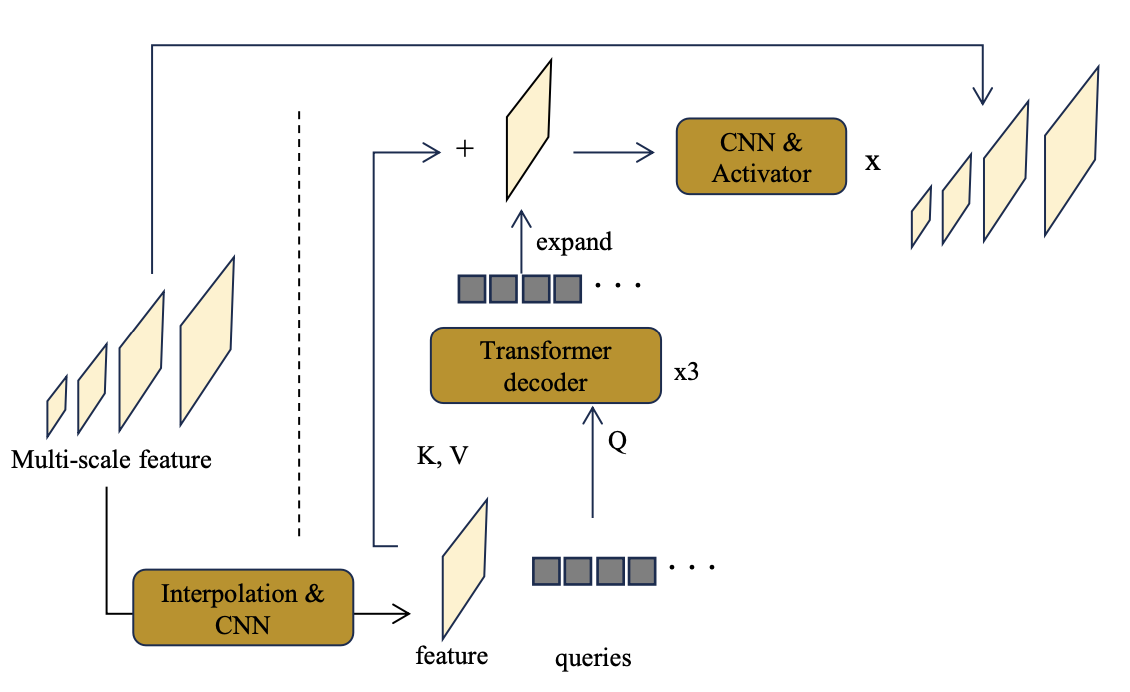}  
    \caption{Region Head overview}  
    \label{Fig:region_head}  
\end{figure}

\noindent  \textbf{Attention-based Region Head}
Our region head enhances traditional convolution-based methods such as MonoDGP\cite{ pu2024monodgp} by integrating query-based Transformer decoders. Initially, multi-scale features are upsampled and concatenated along the channel dimension\cite{liu2023seed}. These fused features are then fed into a Transformer decoder initialized with learnable queries, each independently predicting segmentation masks\cite{antoniussen2024visled}. Ultimately, these query-generated masks are aggregated to form the final segmentation mask, effectively capturing rich contextual information through multi-scale and query-driven processing\cite{doll2022spatialdetr} and the aggregated masks refine object boundaries and provide complementary spatial context. This design plays a crucial role in enhancing the overall 3D detection performance. The detailed architecture of our attention-based region head is illustrated in Figure \ref{Fig:region_head}.

Our attention-based Region Head module augments conventional convolutional approaches by integrating transformer-based decoding. In this module, multi-scale features are first upsampled and concatenated, then processed through a transformer decoder initialized with learnable queries. Each query independently predicts a segmentation mask, and the aggregated masks refine object boundaries and provide complementary spatial context. This design plays a crucial role in enhancing the overall 3D detection performance.

\noindent \textbf{Depth Prediction}
We introduce a learnable depth prediction mechanism that dynamically adjusts depth bin centers based on training data, providing finer granularity and improved regularization. To enhance depth map quality, we integrate a convolution-based refinement module that reduces noise and improves smoothness, supported by an external depth gradient penalty to encourage spatial coherence.We ensure monotonic ordering of depth bin values through a reparameterization strategy where bin intervals are treated as learnable parameters. Given parameters $\delta$ , the bin intervals $\Delta$ are computed as:
\begin{equation}
\Delta_i = \text{softplus}(\delta_i) \cdot \frac{depth_{max} - depth_{min}}{\sum_{j}\text{softplus}(\delta_j)}
\end{equation}
The effective depth bin centers are then calculated as a cumulative sum of these intervals:
\begin{equation}
bin_{k} = depth_{min} + \sum_{i=1}^{k}\Delta_i
\end{equation}
This formulation naturally restricts depth bin centers within the predefined range $[depth_{min},depth_{max}]$, facilitating stable training without additional regularization and ensuring precise control over depth estimation.

We define the region loss as a combination of the Dice loss and Binary Cross Entropy (BCE) loss to enhance segmentation performance across multiple scales. For each scale \(i\) \cite{milletari2016v, taghanaki2019combo}, the Dice coefficient is computed as:
\begin{equation}
D_i = \frac{2\sum (p_i \cdot g_i) + \epsilon}{\sum p_i + \sum g_i + \epsilon},
\end{equation}
where \(p_i\) is the predicted probability map and \(g_i\) is the interpolated ground truth mask. The Dice loss is then defined as \(\mathcal{L}_{Dice,i} = 1 - D_i\) and the BCE loss as:
\begin{equation}
\mathcal{L}_{BCE,i} = -\Bigl[ g_i\log(p_i) + (1 - g_i)\log(1 - p_i) \Bigr].
\end{equation}
The overall region loss is given by
\begin{equation}
\mathcal{L}_{region} = \lambda_{Dice}\frac{1}{N}\sum_{i=1}^{N}\mathcal{L}_{Dice,i} + \lambda_{BCE}\frac{1}{N}\sum_{i=1}^{N}\mathcal{L}_{BCE,i},
\end{equation}
with \(\lambda_{Dice}=0.7\) and \(\lambda_{BCE}=0.3\), effectively combining overlap-based evaluation with pixel-wise classification to boost segmentation performance.

\noindent \textbf{Depth KL Divergence}
To further refine depth predictions, we implement a KL divergence-based regularization between the predicted depth distribution and a sharply peaked target depth distribution. Given predicted depth distributions modeled as Gaussian $p(m,\sigma^{2})$, and target distributions as Gaussian $p(\mu,\epsilon^{2})$ \cite{milletari2016v}, we define the KL divergence regularization loss as follows:
\begin{equation}
 Loss_{Depth\,KL(q||p)} = \log\left(\frac{\sigma}{\epsilon}\right) + \frac{\epsilon^2 + (\mu - m)^2}{2\sigma^2} - \frac{1}{2}
\end{equation}
where $m$ and $\sigma^{2}$ are the predicted depth mean and variance, respectively, $\mu$ is the target depth, and $\epsilon$ is a small hyperparameter defining the sharpness of the target distribution.
\subsection{Virtual Space}
We propose a novel framework called \textit{Virtual Space}, designed to enhance robustness and generalization capabilities in 3D object detection by effectively addressing variability in camera intrinsics and viewpoints\cite{wu2023virtual, brazil2023omni3d}. Extending the concept of virtual depth, Virtual Space introduces additional degrees of freedom that explicitly model camera pose and viewpoint variations\cite{taghanaki2019combo}. Specifically, operations related to extrinsic parameters are performed within a coordinate system centered at the camera used for image acquisition. Given the original image dimensions $(W_{orig}, H_{orig})$ and intrinsic camera parameters, we define a virtual camera characterized by dimensions $(W_{v}, H_{v})$ and focal length $f_{v}$. Scaling factors are subsequently derived based on these defined parameters, facilitating consistent object detection across varying imaging conditions\cite{wu2023virtual}. $s_{x}, x_{y}$ are computed as:
\begin{equation}
s_{x} = \frac{W_{v}}{W_{orig}}, s_{y} = \frac{H_{v}}{H_{orig}}
\end{equation}
\begin{equation}
\begin{bmatrix}
f_{v} & 0 & c_{vx} \\
0 & f_{v} & c_{vy} \\
0 & 0 & 1 \\
\end{bmatrix} 
,\quad where\quad c_{vx}=c_{x} s_{x}, c_{vy} = c_{y} s_{y}
\end{equation}
The transformation from original coordinates  to virtual coordinates  is defined by:
\begin{equation}
    u_v = u \cdot s_x, \quad v_v = v \cdot s_y, \quad Z_v = \frac{f_v}{f_x} Z_{\text{cam}}
\end{equation}
To convert predictions from Virtual Space back to the original camera coordinate system, we apply:
\begin{equation}
\begin{aligned}
    Z_{\text{cam}} &= \frac{f_x}{f_v} Z_v \\
    X_{\text{cam}} &= \frac{(u_v / s_x - c_x) Z_{\text{cam}}}{f_x} \\
    Y_{\text{cam}} &= \frac{(v_v / s_y - c_y) Z_{\text{cam}}}{f_y}
\end{aligned}
\end{equation}
Additionally, within Virtual Space, we introduce viewpoint augmentation and focal length variation to improve robustness.
Training samples are augmented using virtual cameras with multiple focal lengths to generalize depth predictions.
Virtual camera poses are perturbed to achieve viewpoint-invariant representations.
\subsection{Weakly-supervised Learning}
Our framework leverages pseudo-labels generated using foundational models, including a 2D object detector, depth predictor, and orientation predictor \cite{zhou2022cross}. While the outputs from state-of-the-art 2D detectors, 3D predictions derived directly from these outputs often suffer from inaccuracies. To mitigate this, we propose incorporating a consistency loss that aligns 2D predictions with corresponding 3D estimations \cite{liu2022learning}.

\noindent \textbf{Pseudo-label Generation} The pseudo-label generation involves several stages. First, Detect objects using 2D detector, obtaining 2D bounding boxes and confidence scores \cite{ren2015faster}. For each detected bounding box, we inverse-project the central point of the 2D bounding box to virtual 3D space using the depth prediction \cite{ranftl2022towards}. If the projected center point falls within another detection bounding box, we select an alternative point located within the central quarter of the bounding box to avoid conflicts. We acquire object orientation using a foundational orientation prediction model \cite{mousavian20173d}. Object dimensions are heuristically estimated from the detected 2D bounding box dimensions and orientation \cite{brazil2019m3d}.
\begin{equation}
\begin{aligned}
h &= \frac{(\text{bottom} - \text{top}) \times z}{f_y}, \\
\text{width}_{2D} &= \frac{f_x}{z} \Bigl(|w_{\text{base}} \cos(\text{yaw})| + |l_{\text{base}} \sin(\text{yaw})|\Bigr), \\
\text{scale} &= \left| \frac{\text{right} - \text{left}}{\text{width}_{2D}} \right|, \\
w &= w_{\text{base}} \times \left[ \text{scale} \right]_{\alpha}^{\beta}, \\
l &= l_{\text{base}} \times \left[ \text{scale} \right]_{\alpha}^{\beta}.
\end{aligned}
\end{equation}

Our dataset, Synergy3D, was created using this pseudo-label generation approach, eliminating the need for human labeling efforts.

\noindent \textbf{Consistency Loss}
Given the generally higher accuracy of 2D detection compared to 3D predictions, we propose a consistency loss designed to align the dimensions of 2D bounding boxes with their corresponding projections derived from the estimated 3D sizes and predicted yaw angles. This loss is applied solely during the weak supervision phase, with the 2D predictions frozen to preserve their accuracy. Through the introduction of this consistency loss, our method effectively leverages relatively noisy 3D pseudo ground truths for robust training, underscoring the importance of this alignment step in facilitating successful 3D detection using pseudo labels.

\begin{equation}
\text{s}_{\text{proj}} = \frac{f_x \times Dim_{3D}}{Depth}
\end{equation}

\begin{equation}
L_{\text{consistency}} = \text{Smooth-}L_1\!\Big( \big[\,s_{\text{proj}} - s_{2D}\,\big]_{-{\beta}}^{+{\beta}}\Big)
\end{equation}


\noindent \textbf{Regularization}
To improve robustness during weakly-supervised learning, we introduce two regularization strategies: statistical outlier rejection and L2 regularization \cite{zhou2019robust}.
We dynamically filter outlier predictions based on the deviation of their losses \cite{huber1992robust}. Specifically, we calculate the median and standard deviation of the losses across predictions, establishing a dynamic threshold. Predictions with losses exceeding this threshold are considered outliers and discarded \cite{rousseeuw2005robust}.


\begin{equation}
\tau = \text{median}(\{L_j\}) + k \cdot \text{std}(\{L_j\})
\end{equation}
\[
L_i' = 
\begin{cases}
L_i, & \text{if } L_i \leq \tau \\[6pt]
0, & \text{otherwise (exclude or down-weight)}
\end{cases}
\]

Additionally, we apply a weighted L2 regularization term to the model parameters to prevent overfitting and enhance generalization \cite{goodfellow2016deep}:

\begin{equation}
L_{reg} = \lambda \sum_{\theta \in \Theta} |\theta|^2
\end{equation}

where $\lambda$ is a hyperparameter controlling the strength of regularization, and $\Theta$ denotes the model parameters. This regularization helps stabilize training and improves model performance.

%% file: sec/4_experiments.tex
\section{Experiments}
\label{sec:experiments}

\begin{table}[]
\centering
\resizebox{\linewidth}{!}{%
\begin{tabular}{c|c|ccc} 
\hline
\multirow{2}{*}{\textbf{Model}} & \multirow{2}{*}{\textbf{Extra Data}} & \multicolumn{3}{c}{$AP_{3D}, IOU=0.7|R_{40}$} \\ \cline{3-5} 
 &  & Easy & Mod & Hard \\ \hline 
CaDDN & \multirow{4}{*}{Lidar}  & 19.17\% & 13.41\% & 11.46\%  \\ 
MonoDTR &  & 21.99\% & 15.39\% & 12.73\% \\ 
DID-M3D &  & 24.40\% & 16.29\% & 13.75\% \\ 
OccupancyM3D &  & 25.55\% & 17.02\% & 14.79\%  \\ \hline
MonoPGC & \multirow{2}{*}{Depth} & 24.68\% & 17.17\% & 14.14\% \\ 
OPA-3D &  & 24.60\% & 17.05\% & 14.25\% \\ \hline
GUPNet & \multirow{10}{*}{None} & 20.11\% & 14.20\% & 11.77\% \\ 
MonoCon &  & 22.50\% &16.46\%& 13.95\% \\
DEVIANT &  & 21.88\% &14.46\%& 11.89\% \\
MonoDDE &  & 24.93\% &17.14\%& 15.10\% \\
CubeRCNN &  & 23.59\% & 15.01\% & 12.56\% \\
MonoUNI & & 24.75\% &16.73\%& 13.49\% \\
MonoDETR &  & 25.00\% &16.47\%&13.58\%\\
MonoCD &  & 25.53\% &16.59\% &14.53\% \\
FD3D &  & 25.38\%& 17.12\%& 14.50\% \\
MonoDGP &  & \textbf{26.35\%} &\underline{18.72\%} &\underline{15.97\%} \\ \hline
Ours & None & \underline{26.07\%} & \textbf{18.85\%} & \textbf{16.76\%} \\ 
Improvement & - & \color[HTML]{3166FF}-0.28\% & \color[HTML]{FE0000}+0.13\% & \color[HTML]{FE0000}+0.79\% \\ \hline
\end{tabular}
}
\caption{KITTI Test Car}
\label{t:kitti_test_car}
\end{table}

\begin{table}[]
\centering
\resizebox{\linewidth}{!}{%
\begin{tabular}{c|c|ccc} 
\hline
\multirow{2}{*}{\textbf{Model}} & \multirow{2}{*}{\textbf{Extra Data}} & \multicolumn{3}{c}{$AP_{3D}, IOU=0.5|R_{40}$} \\ \cline{3-5} 
 &  & Easy & Mod & Hard \\ \hline 
CaDDN & \multirow{2}{*}{Lidar}  & 12.87\% &8.14\% &6.76\%  \\ 
OccupancyM3D &  & 14.68 \%&9.15 \%&7.80\%  \\ \hline
MonoPGC & \multirow{1}{*}{Depth} & 14.16 \%&9.67\%& 8.26\% \\ \hline
GUPNet & \multirow{6}{*}{None} & 14.72\%& 9.53\% &7.87\% \\ 
MonoCon &  & 13.10\% &8.41\%& 6.94\% \\
DEVIANT &  & 13.43\% & 8.65\%& 7.69\% \\
MonoDDE &  & 11.13\% &7.32\%& 6.67 \%\\
MonoDETR &  & 12.65\% & 7.19\% &6.72\%\\
MonoDGP &  & \underline{15.04\%} & \underline{9.89\%}& \underline{8.38\%} \\ \hline
Ours & None & \textbf{16.25\%} & \textbf{10.53\%} & \textbf{8.91\%} \\
Improvement & - & \color[HTML]{FE0000}+1.21\% & \color[HTML]{FE0000}+0.64\% & \color[HTML]{FE0000}+0.53\% \\ \hline
\end{tabular}
}
\caption{KITTI Test Pedestrian}
\label{t:kitti_test_ped}
\end{table}


\subsection{Datasets}
We utilize widely recognized benchmark datasets, including KITTI\cite{geiger2012kitti}, Waymo\cite{sun2020waymo}, and nuScenes\cite{caesar2020nuscenes} (with GT), to pretrain our model. Additionally, we introduce a novel dataset, Synergy3D (with pseudo GT), specifically designed to enhance generalization capabilities. Furthermore, we employ our collected office dataset (with GT) to evaluate the generalization performance of our approach.

\noindent \textbf{KITTI} dataset consists of 7,481 training images and 7,518 testing images, categorized into three classes: \textit{Car, Pedestrian, and Cyclist}. Each class is further divided into three difficulty levels (\textit{Easy, Moderate, and Hard}). We split the 7,481 training images into a training set of 3,712 images and a validation set of 3,769 images for our ablation study. In accordance with the official evaluation protocol, we use \textit{AP3D | R40} and \textit{APBEV | R40} on the \textit{Moderate} category as our primary evaluation metrics.

\noindent \textbf{Synergy3D Dataset} The proposed Synergy3D dataset is pivotal to our framework, as it drastically reduces the labeling cost compared to traditional methods. Unlike prior works such as Omni3D\cite{brazil2023omni3d}, Synergy3D covers a broader range of environments (including indoor, shopping malls and public spaces) while maintaining a labeling cost that is orders of magnitude lower. As demonstrated by the significant improvements reported in \ref{t:office}, the diversity of Synergy3D, when combined with our weakly-supervised learning strategy, contributes synergistically to the performance gains of GATE3D. This synergy between dataset diversity and architectural innovations underscores the potential of our approach for generalized 3D object detection in real-world applications.

\noindent \textbf{Office Dataset.} The Office Dataset consists of synchronized LiDAR point clouds and camera-derived 3D data. We manually annotated 3D labels directly on the LiDAR points and subsequently transformed these annotations into camera coordinates. The dataset includes a single class: person (encompassing sitting, walking, and eating behaviors). It is specifically designed to benchmark and evaluate our model's generalization capabilities in previously unseen domain environments.

\begin{table*}[] 
\centering
\begin{tabular}{c|c|c|c|c|c|c} 
\hline
\multicolumn{1}{c|}{\textbf{Query Diversity}} & 
\multicolumn{1}{c|}{\textbf{Adaptive Fusion}} & 
\multicolumn{1}{c|}{\textbf{Region Loss}} & 
\multicolumn{1}{c|}{\textbf{Consistency Loss}} & 
\multicolumn{1}{c|}{\textbf{$AP_{3D}$}} & 
\multicolumn{1}{c|}{\textbf{$AP_{BEV}$}} & 
\multicolumn{1}{c}{\textbf{$AP_{bbox}$}} \\ \hline
\ding{55}  & \ding{55}  & \ding{55}  & \ding{55}  & 10.49\%  & 11.76\%  & 82.76\%\\
\ding{51}  & \ding{55}  & \ding{55}  & \ding{55} & 11.01\%  & 12.18\% & 83.16\% \\
\ding{51}  & \ding{51}  & \ding{55}  & \ding{55}  & 12.32\%  & 13.23\% & 83.76\% \\
\ding{51}  & \ding{51}  & \ding{51}  & \ding{51} & 14.83\%  & 15.88\%  & 84.39\%\\ \hline
\multicolumn{4}{c|}{\textbf{Improvement}}  & \color[HTML]{FE0000} +4.34\%  & \color[HTML]{FE0000} +4.12\%  & \color[HTML]{FE0000}+1.63\%\\ \hline
\end{tabular}
\caption{Ablation Study}
\label{t:ablation}
\end{table*} 

\subsection{Results}
\noindent \textbf{Model Performance} To thoroughly assess the effectiveness of our proposed GATE3D model, we trained exclusively on the KITTI dataset and compared its performance against multiple state-of-the-art methods that employ various additional input modalities such as lidar and depth, as well as purely monocular models. As detailed in Tables~\ref{t:kitti_test_car} and \ref{t:kitti_test_ped}, GATE3D demonstrates highly competitive results across all evaluation metrics.
In the car detection task (Table~\ref{t:kitti_test_car}), our model achieves the best performance among monocular approaches in the moderate (18.85\%) and hard (16.76\%) difficulty categories, surpassing existing methods such as MonoDGP\cite{pu2024monodgp}, FD3D\cite{wu2024fd3d}, and MonoCD\cite{yan2024monocd}. Specifically, GATE3D significantly enhances detection accuracy in challenging scenarios, reflecting its robustness and effectiveness under complex visual conditions typical of real-world applications.
Furthermore, in the pedestrian detection task (Table~\ref{t:kitti_test_ped}), our GATE3D model outperforms all compared models—including those leveraging extra modalities like lidar and depth—across easy (16.25\%), moderate (10.53\%), and hard (8.91\%) difficulty settings.
Overall, these results confirm the superior performance and practical potential of our GATE3D model in monocular 3D detection tasks, highlighting its strength in handling challenging environmental conditions without relying on additional sensor inputs.

\noindent \textbf{Generalization Performance} To rigorously evaluate the generalization capability of our proposed GATE3D model, we tested its performance on a completely different and previously unseen domain—the Office Dataset. Notably, this dataset was never included in the training procedure, ensuring a robust assessment of the model's domain generalization capabilities. As presented in Table~\ref{t:office}, our GATE3D model, trained on the combined $Synergy3D^{*}$ dataset (which integrates KITTI, nuScenes, Waymo, and Synergy3D datasets using our weakly supervised learning framework), significantly surpasses existing state-of-the-art models. Specifically, GATE3D demonstrates substantial improvements across all evaluated metrics, including $AP_{3D}$ (14.83\%, +4.4\%), $AP_{BEV}$ (15.88\%, +4.76\%), and $AP_{bbox}$ (84.39\%, +13.89\%).
For context, K+N+W denotes models trained on the combined KITTI, nuScenes, and Waymo datasets, while Omni3D\cite{brazil2023omni3d} combines diverse datasets such as SUN RGB-D\cite{song2015sun}, ARKitScenes\cite{dehghan2021arkitscenes}, Hypersim\cite{roberts2021hypersim}, Objectron\cite{ahmadyan2021objectron}, KITTI\cite{{geiger2012kitti}}, and nuScenes\cite{caesar2020nuscenes}. The remarkable performance of GATE3D in the Office Dataset highlights its robust ability to generalize beyond training domains, confirming its practical utility and effectiveness in handling diverse and real-world scenarios.

\begin{table}[]
\centering
\caption{People in Office}
\label{t:office}
\resizebox{\linewidth}{!}{%
\begin{tabular}{l|l|lll}
\hline
\multicolumn{1}{c|}{\multirow{2}{*}{\textbf{Model}}} & \multicolumn{1}{c|}{\multirow{2}{*}{\textbf{Dataset}}} & \multicolumn{3}{c}{$AP_{3D}, IOU=0.5|R_{40}$} \\ \cline{3-5} 
\multicolumn{1}{c|}{} & \multicolumn{1}{c|}{} & \multicolumn{1}{c}{$AP_{3D}$} & \multicolumn{1}{c}{$AP_{BEV}$} & \multicolumn{1}{c}{$AP_{bbox}$} \\ \hline
DEVIANT & K+N+W & 0.81\% & 1.23\% & 5.12\%  \\ 
MonoDETR & K+N+W & 0.81\% & 1.33\% & 5.45\% \\ 
MonoDGP & K+N+W & 0.88\% & 1.85\% & 5.50\% \\ \hline
CubeRCNN & Omni3D & 1.55\% & 3.35\% & 8.60\% \\
MonoDGP & $Synergy3D^{*}$ & \underline{10.43\%} & \underline{11.12\%} & \underline{70.50\% }\\ \hline
Ours & $Synergy3D^{*}$ & \textbf{14.83\%} & \textbf{15.88\%} & \textbf{84.39\%} \\ 
Improvement & - & \color[HTML]{FE0000}+4.4\% & \color[HTML]{FE0000}+4.76\% & \color[HTML]{FE0000}+13.89\% \\ \hline
\end{tabular}%
}
\end{table}

\subsection{Ablation Studies}
To identify key factors contributing to domain generalization performance, we conducted detailed ablation studies using our novel Office dataset, a completely different domain not included during training.

\noindent \textbf{Query Diversity}
As shown in Table~\ref{t:ablation}, introducing Query Diversity significantly improves detection accuracy, raising the $AP_{3D}$ score from 10.49\% to 11.01\%. By enforcing reduced cosine similarity among query vectors, this component minimizes duplicate predictions, leading to increased stability and precision in detection.

\noindent \textbf{Adaptive Fusion}
Integration of the Adaptive Fusion module further enhances model performance, increasing $AP_{3D}$ from 11.01\% to 12.32\%. This improvement highlights the effectiveness of the adaptive gating mechanism, dynamically incorporating multi-scale and depth-aware features, thus providing richer contextual information for more accurate 3D object detection.

\noindent \textbf{Region Loss}
Adding Region Loss, which combines Dice and Binary Cross-Entropy (BCE) losses, contributes notably to further performance improvements. Specifically, AP$_{3D}$ rises from 12.32\% to 14.83\%, indicating that the Region Loss significantly boosts segmentation accuracy and facilitates finer-grained predictions.

\noindent \textbf{Consistency Loss}
The introduction of Consistency Loss demonstrates substantial effectiveness, particularly addressing label noise in 3D annotations. This component further solidifies the model's predictive consistency, enhancing robustness against inaccurate or noisy labels, and contributes significantly to the final $AP_{3D}$ performance (14.83\%). Overall, the comprehensive inclusion of these four components yields cumulative improvements of +4.34\%, +4.12\%, and +1.63\% in $AP_{3D}$, $AP_{BEV}$, and $AP_{bbox}$, respectively.


%% file: sec/5_conclusion.tex
\section{Conclusion}
In this work, we evaluated the generalization capabilities of our proposed GATE3D framework on critical object classes (i.e., persons and cars), addressing the challenge of limited benchmarks for diverse 3D detection scenarios. Our experimental results demonstrate that GATE3D robustly generalizes across different environments, thereby providing a solid foundation for applications in augmented and mixed reality, where accurate scene understanding is essential. Future research will focus on extending our framework to additional object categories and incorporating comprehensive orientation estimation—including roll and pitch angles—to further enhance detection performance in complex, interactive environments.

\section{Appendix}
\begin{figure}[htbp]
    \centering
    \includegraphics[width=\columnwidth]{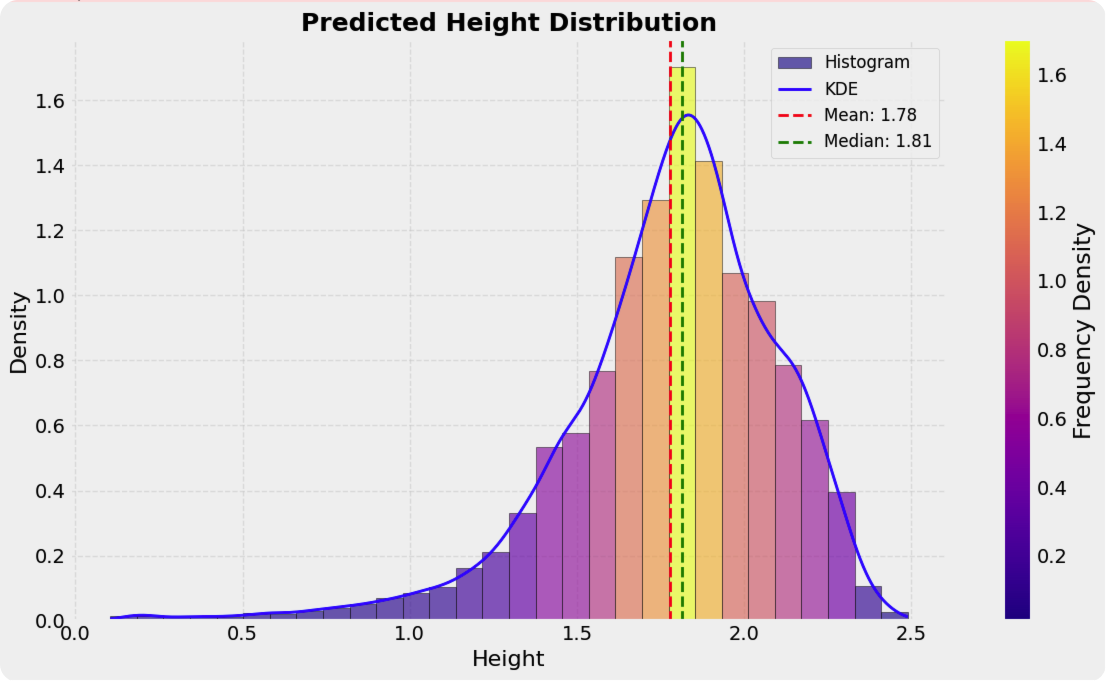}  
    \caption{ Distribution of predicted person heights. This histogram shows the model’s predicted heights for detected people in the evaluation set. The distribution is centered around the average adult height (~1.7 m), indicating that most detections have realistic scales. A smaller secondary peak at lower heights suggests instances of seated individuals or partial detections, demonstrating that GATE3D’s outputs remain physically plausible.
}  
    \label{Fig:hist}  
\end{figure}
\noindent \textbf{Implementation Details}
Our model utilizes Pyramid Vision Transformer (PVT) as the backbone architecture, where each Transformer attention module comprises 8 attention heads and employs 50 object queries. Multi-scale deformable attention mechanisms leverage 4 sampling points. For the virtual camera configuration, we set the focal length to 900 with an image resolution of 1274×644 pixels. During inference, queries exhibiting category confidence scores below 0.1 are eliminated, and outlier predictions are filtered based on a threshold of 2.0 standard deviations established during weakly supervised training. The Synergy3D dataset used for training contains approximately 100K pseudo-labeled samples. Notably, our inference pipeline employs a category confidence threshold without applying Non-Maximum Suppression (NMS). Training was conducted on four NVIDIA A6000 GPUs, achieving an average inference speed of 82ms per image at an input resolution of 1274x644 pixels, utilizing approximately 3.1GB of GPU memory.


\noindent \textbf{Qualitative Results}
Figure \ref{Fig:q_test} presents qualitative results illustrating predicted 3D information projected onto 2D images. These evaluations utilize the Object365\cite{ahmadyan2021objectron} and MOT datasets\cite{voigtlaender2019mots}, which were not part of the training dataset. Figure \ref{Fig:hist} shows the height distribution of detected persons, illustrating that predictions generally match average human heights, with smaller detections potentially corresponding to seated individuals.
Figure \ref{fig:height_accuracy} further combines qualitative frames with a per‑frame height trace, demonstrating that the predicted boxes maintain physically consistent scale throughout the unseen indoor sequence.

%% file: sec/X_suppl.tex
\clearpage
\setcounter{page}{1}


\begin{figure}[htbp]
    \centering
    \includegraphics[width=\textwidth, height=0.90\textheight]{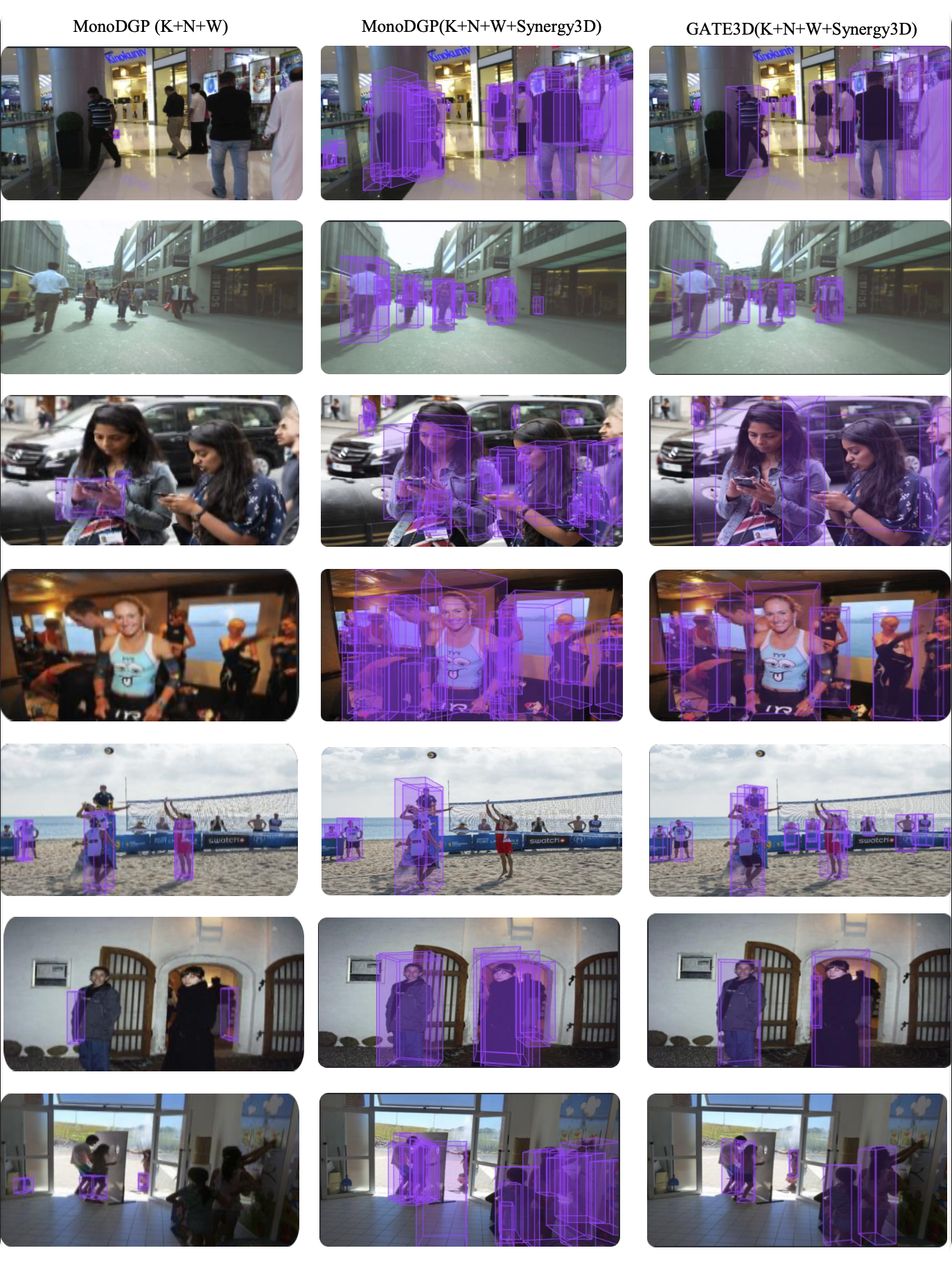}
    \caption{Qualitative 3D detection results on unseen datasets}
    \label{Fig:q_test}
\end{figure}







\begin{figure*}[htbp]
  \centering
  \begin{subfigure}[t]{\linewidth}
      \centering
      \includegraphics[width=\linewidth,keepaspectratio]{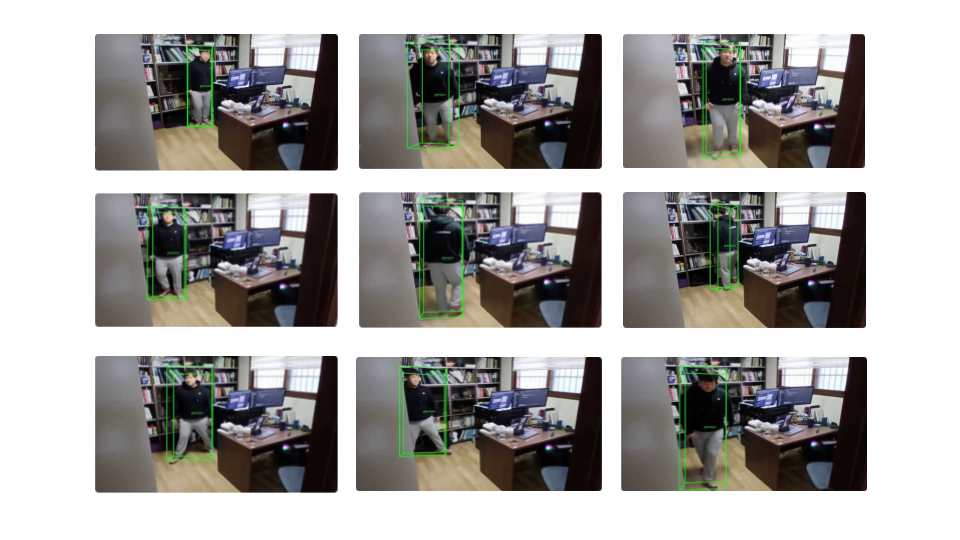}
      \label{fig:home_pic}
  \end{subfigure}
  
  \vspace{0.8em}  
  
  \begin{subfigure}[t]{\linewidth}
      \centering
      \includegraphics[width=\linewidth,keepaspectratio]{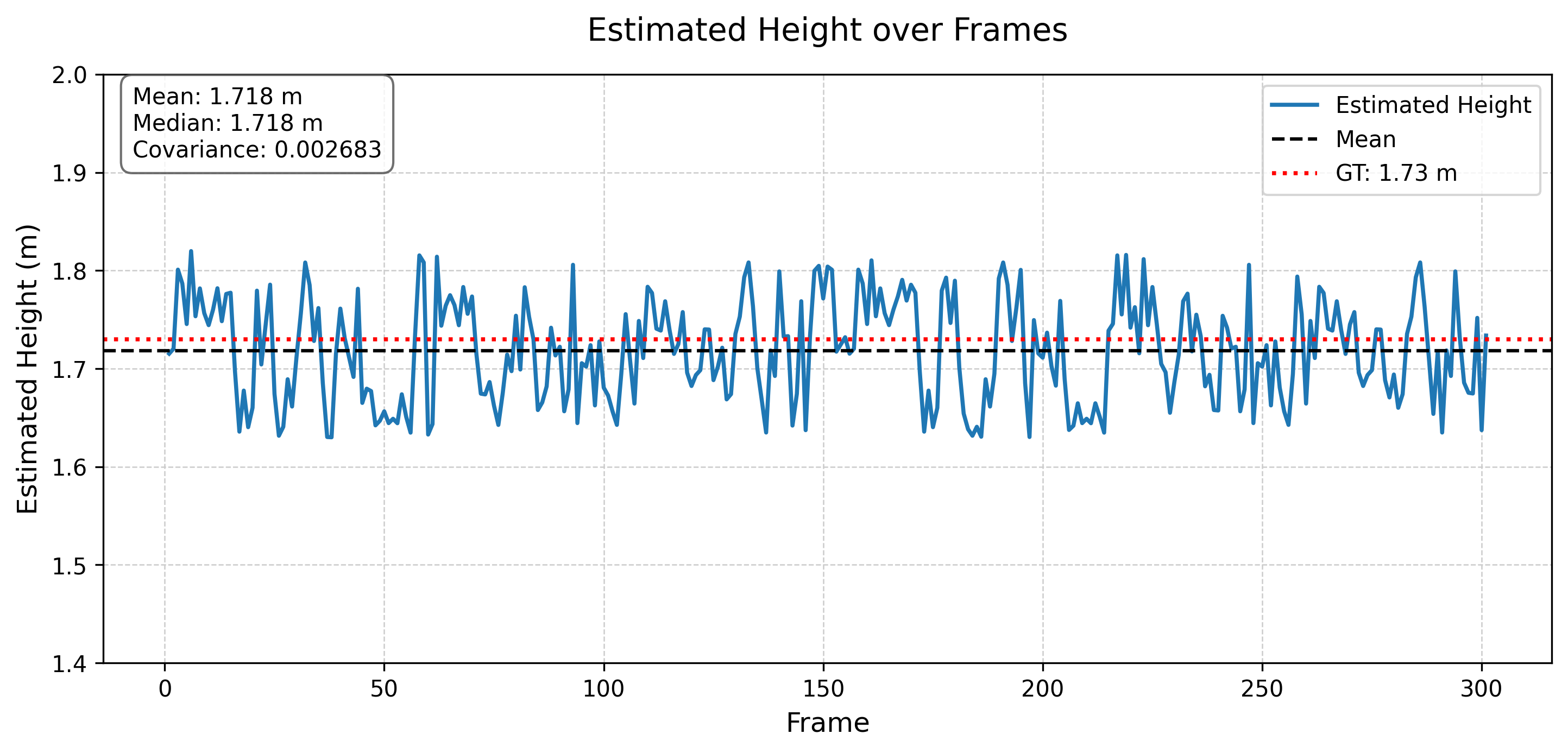}
      \label{fig:height_plot}
  \end{subfigure}
  
  \caption{\textbf{Metric‑scale fidelity in a novel environment.}
  The qualitative snapshots and the quantitative height trace jointly demonstrate that \emph{GATE3D} retains physically consistent scale when applied to an unseen indoor scene, confirming its reliability for downstream mixed‑reality applications. Per‑frame height estimates for the same sequence. The subject’s ground‑truth stature is \textbf{1.73m}, whereas GATE3D yields
      mean =\textbf{1.718m}, median=\textbf{1.718m}, and variance $=2.683\times10^{-3}\,\mathrm{m}^{2}$
\;($\sigma\!\approx\!5.18\,\mathrm{cm}$) across 300 frames.}
  \label{fig:height_accuracy}
\end{figure*}